\documentclass[11pt]{article}
\usepackage{authblk}
\usepackage{coling2020}
\usepackage{times}
\usepackage{url}
\usepackage{latexsym}
\usepackage{amsmath}
\usepackage{xcolor}
\usepackage{graphicx}
\usepackage{appendix}
\usepackage{listings}
\usepackage{xspace}
\usepackage{float}
\usepackage{enumitem}
\usepackage{amssymb}
\usepackage{makecell}
\usepackage{xcolor}
\usepackage[font=small,labelfont=bf]{caption}
\newcommand\ab[1]{}
\newcommand\edit[1]{{\color{black}{#1} }}
\renewcommand{\emph}[1]{\textbf{#1}}
\newcommand{\name}{DILOG\xspace}

\colingfinalcopy % Uncomment this line for the final submission

\title{Resource Constrained Dialog Policy Learning \\via Differentiable Inductive Logic Programming}
\author{Zhenpeng Zhou}
\author{Ahmad Beirami}
\author{Paul Crook}
\author{Pararth Shah}
\author{\\Rajen Subba}
\author{Alborz Geramifard}
\affil{Facebook}
\affil[ ]{\tt \normalsize \{zzp, beirami, pacrook, pararths, rasubba, alborzg\}@fb.com}
\date{}

\begin{document}

\setlength{\abovedisplayskip}{4pt}
\setlength{\belowdisplayskip}{4pt}

\maketitle

\begin{abstract}
  Motivated by the needs of resource constrained dialog policy learning,
  we introduce dialog policy via differentiable inductive logic (\name). We explore the tasks of one-shot learning and zero-shot domain transfer with \name on SimDial and MultiWoZ. Using a {\it single} representative dialog from the restaurant domain, we train \name on the SimDial dataset and obtain 99+\% in-domain test accuracy. We also show that the trained DILOG zero-shot transfers to all other domains with 99+\% accuracy, proving the suitability of \name to slot-filling dialogs. We further extend our study to the MultiWoZ dataset achieving 90+\% inform and success metrics. We also observe that these metrics are not capturing some of the shortcomings of DILOG in terms of false positives, prompting us to measure an auxiliary Action F1 score. We show that \name is 100x more data efficient than state-of-the-art neural approaches on MultiWoZ while achieving similar performance metrics. We conclude with a discussion on the strengths and weaknesses of \name.
\end{abstract}

\section{Introduction}

\blfootnote{
    \hspace{-0.65cm}  % space normally used by the marker
    This work is licensed under a Creative Commons 
    Attribution 4.0 International License.
    License details:
    \url{http://creativecommons.org/licenses/by/4.0/}.
}
To foster research on dialog policy learning for virtual digital assistants, several task-oriented dialog corpora have been introduced in recent years, such as SimDial~\cite{zhao2018zero}, MultiWoZ~\cite{budzianowski2018multiwoz}, Taskmaster~\cite{Taskmaster1}, and Schema Guided Dialog~\cite{rastogi2019towards}, to name a few. 
Deep learning approaches, including mixture models~\cite{pei2019retrospective} hierarchical encoder/decoder~\cite{zhang2019task,chen2019semantically}, reinforcement learning~\cite{zhao2019rethinking}, and pre-trained language models~\cite{wu2019alternating,peng2020soloist,hosseini2020simple}, have significantly advanced dialog policy research in the past few years~\cite{gao2019neural}, setting new state-of-the-art performance limits. 
%More recently, SimpleTOD~\cite{hosseini2020simple} and SOLOIST~\cite{peng2020soloist} have shown that pre-training dialog policy using large language models, e.g., GPT-2~\cite{radford2019language}, can lead to significantly better performance on task-oriented neural dialog policy learning by using even larger neural models.

%\ab{somewhere we want to mention that data collection is expensive -- both in time; and other resources}

However, collecting annotated data for supervised dialog policy learning is an expensive and time-consuming process. Hence, it is desirable to explore approaches to train dialog policy with limited data and transfer an existing policy with few or even no additional training data to new domains. 
This practical requirement has motivated the community to research resource-constrained dialog policy learning in the past few decades. Researchers have explored approaches including employing grammar constraints for dialog policy~\cite{eshghi2017bootstrapping}, transfer learning \cite{shalyminov2019few}, or pre-trained language models~\cite{zhao2020low}. Few-shot domain adaptation has been researched since the 2000s~\cite{litman2002designing} on both end-to-end dialog systems \cite{qian2019domain,zhao2018zero} as well as dialog policy learning~\cite{vlasov2018few}.

\edit {In a traditional modular dialog system, the dialog policy aims to decide a dialog action given a dialog state, while assuming the tasks of language understanding and generation are handled by other components. Under such assumptions, a task-oriented dialog policy mostly follow the slot-filling scheme, which can be described by a set of probabilistic rules. Therefore, we hypothesize that dialog policy in this limited sense can be constructed by learning the underlying rules. To this end, we draw upon the recent advances in developing differentiable inductive logical programs (DILP) \cite{evans2018learning} that use neural architectures to learn almost rule-based policies.} We present \name, an adaptation of DILP  to dialog policy learning. Briefly, \name discerns a set of logical rules from the examples by using inductive reasoning.\footnote{Inductive reasoning tries to summarize general principles from special cases. For example, the fact ``cars A, B, and C drive on the right side of the road'' \emph{induces} that ``all cars drive on the right side of the road.''}  We introduce \name in Section~\ref{sec:DILP}.
We apply \name to the SimDial Dataset~\cite{zhao2018zero} (Section~\ref{sec:simdial-zeroshot}), and MultiWoZ Dataset~\cite{budzianowski2018multiwoz}
(Section~\ref{sec:multiwoz-zeroshot}), showing that on the task of one-shot dialog policy learning and zero-shot domain transfer, \name outperforms several other neural baselines.
Finally, Section~\ref{sec:conclusion} concludes this paper.

\section{DILOG: Dialog Policy Via Differentiable Inductive Logic} 
\label{sec:DILP}
Inductive Logic Programming (ILP) is a paradigm which derives a hypothesized first-order logic given the background knowledge, positive, and negative examples~\cite{muggleton1994inductive}. The central component of ILP is known as clauses. A \emph{clause} is a rule expressed as $\alpha \leftarrow \alpha_1,..., \alpha_n$, where $\alpha$ is the \emph{head} atom and $\alpha_1,..., \alpha_n$ are \emph{body} atoms. An \emph{atom} $p(t_1,...,t_m)$ is composed of an $m$-ary \emph{predicate} $p$ and a tuple of \emph{terms} $t_1,...,t_m$, which can be variables or constants. An atom is \emph{ground} if it only contains constants. For example, The following clause defines when to perform the action of confirmation:
$$\mathtt{confirm}(S) \leftarrow \mathtt{user\_request}(S, T), \mathtt{not\_confident}(S),$$
which means if the user requests a slot $S$ in a task $T$, and the system is not confident about $S$, then the system should confirm $S$ with user. Employing the clause above on the grounding atoms of $\mathtt{user\_request}(contact, calling)$ and $\mathtt{not\_confident}(contact)$, we will be able to deduce the  action of $\mathtt{confirm}(contact)$.

DILP~\cite{evans2018learning} combines  ILP with a differentiable neural network architecture, to make ILP robust to noisy or ambiguous data. In short, DILP generates a collection of clauses based on the rule templates,\footnote{A template describes the rule used to generate the clauses. \edit{A rule template is defined by the number of additional variables $v\in\mathbb{N} $ and whether to allow intensional predicates $i\in \{0, 1\}$. For example, we can specify the template for predicate $\mathtt{p}(X)$ to be $\{v=0, i=1\}$, which means using no additional variables (only $X$), and allowing intensional predicates. When two other 1-ary predicates exist as \texttt{q} and \texttt{r}, we first generate all possible atoms, namely $\mathtt{q}(X)$, and $\mathtt{r}(X)$, then we enumarate all combination of those atoms to generate the body of the clause.} Therefore the clauses generated will be $\mathtt{p}(X)\leftarrow \mathtt{q}(X),\mathtt{q}(X)$, $\mathtt{p}(X)\leftarrow \mathtt{r}(X),\mathtt{r}(X)$, and $\mathtt{p}(X)\leftarrow \mathtt{q}(X), \mathtt{r}(X)$. \edit{We refer to \cite{evans2018learning} for a detailed elaboration on the generation process, and Appendix~\ref{app:progtemp} for the process of choosing hyperparameters in a rule template.}} and assigns trainable weights to those clauses. Then logical deduction is applied recursively using the weighted sum of the clauses on the valuation vector $\mathbf{a}\in [0, 1]^g$, where $g$ is the number of grounding atoms and $\mathbf{a}[i]$ is the probability that the grounding atom $i$ is true. With the trainable weights, DILP can be trained using the gradient descent method.

\name is based on DILP with several modifications: (1) We include an option of adding $\ell_1$ or $\ell_2$ regularizers to the weight matrix of the clauses to improve generalization. (2) We allow adding clauses as background knowledge, so as to enable continual learning using pre-learned rules. (3) We use element-wise max instead of probabilistic sum as the amalgamate function to update the valuation vector. As probabilistic sum is accumulating the valuation at each step, it is easier to get into a local optimum when the inference steps are long. (4) In DILP, a problem is defined as $(\mathcal{L}, \mathcal{B}, \mathcal{P}, \mathcal{N})$, where $\mathcal{L}$ is the language frame to generate potential clauses, $\mathcal{B}$  is the background knowledge, $\mathcal{P}$ is the positive examples, and $\mathcal{N}$ is the negative examples. In this definition, all positive and negative examples are grounded on the same set of constants $\mathcal{C}$. As the number of clauses scales quadratically with the number of constants, it can be computationally expensive when the sample size is large. Comparatively, we define a problem as $(\mathcal{L}, \mathcal{S})$, where $\mathcal{S}$ is the set of samples, and each $x\in \mathcal{S}$ is a tuple of $(\mathcal{B}, \mathcal{P}, \mathcal{N}, \mathcal{C})$. This modification allows each sample to define its own set of constants, which makes it computationally tractable under the setting of dialog system where each dialog consists of multiple turns.

In this paper, we demonstrate the application of \name on two tasks in dialog policy learning: one-shot learning and zero-shot domain transfer. In a modular dialog system, which consists of automatic speech recognition, natural language understanding, dialog state tracking (DST), dialog manager, and natural language generation (NLG), the role of dialog policy is to map every state $s\in\mathbb{S}$ (represented by the DST) to a dialog action $a\in\mathbb{A}$. Compared with end-to-end policy learning that generates a natural language response, we only focus on learning the policy function $\pi: \mathbb{S} \rightarrow \mathbb{A}$, and leave NLG as a separate problem.  Under this setting, \name offers several advantages: 
\begin{itemize}[topsep=2pt,itemsep=2pt,partopsep=2pt, parsep=1pt]
    \item {\bf Sample efficiency}: \name generalizes well from a small number of samples by introducing a language bias from the template used to generate the clauses, with which a set of succinct rule is preferred over a set of complex rules. This makes it useful in one-shot policy learning.
    
    \item {\bf Interpretability}: The dialog policy learned via \name consists of a set of (probabilistic) rules. Each of the rules can be manually inspected and understood. This feature is desirable in industrial settings where interpretability and debuggability are key considerations.
    
    \item {\bf Domain generalizability}: The rules learned by \name in one domain can be zero-shot transferred to new domains with unseen slots, by assuming symmetry between slots. This helps quickly enable a new domain with no new data collection and annotation.

\end{itemize}
%We will demonstrate those advantages on the policy learning tasks on the SimDial and MultiWoZ datasets.

\section{\name on the SimDial Dataset}
\label{sec:simdial-zeroshot}

SimDial~\cite{zhao2018zero} is a multi-domain dialog generator that can generate conversations in domains including restaurant, movie, bus, and weather. Each domain is defined by a collection of meta data including user slots and systems slots. Ignoring the actions of greeting and goodbye, possible user actions are inform or request, while system actions include inform, request, and database query. 
We use SimDial to generate clean dialogs from all four domains, among which a single {\it representative} dialog (that contains all user and system actions) from the restaurant domain is used for training and 500 dialogs from each other domains are used as the test set.

\begin{table}[h]
\edit{
\caption{An illustrative example of adapting the dialog state and actions to the logical forms of \name. Note that we use a linked list to enumerate the user slots in the logical form, where $\mathtt{succ}(X, Y)$ denotes the successor of $X$ is $Y$, and $\mathtt{terminal}(X)$ means $X$ is a terminal node. The first 4 predicates of the belief state describe a linked list of [usr\_slot] $\rightarrow$ [food\_pref] $\rightarrow$ [loc] $\rightarrow$ [term].}
\label{tab:illus}
\center
\begin{minipage}{0.35\textwidth}
    Dialog:\\

    \resizebox{\textwidth}{!}{
    \hspace{1em}\makecell[l]{
        Previous Turn(s):\\
        \hspace{1em}{\color{gray}USR: What’s up? I need a restaurant.} \\ 
        \hspace{1em}{\color{gray}SYS: Which place?} \\
        Current Turn:\\
        \hspace{1em}USR: I am at San Jose. \\ 
        \hspace{1em}SYS: What kind of food do you like?}

    }
\end{minipage}
\hspace{0.5em}
\begin{minipage}{0.6\textwidth}
\resizebox{\textwidth}{!}{
\begin{tabular}{c|c|c}
    \hline
    & State Representation & Logical Form \\
    \hline
    \makecell[l]{Belief \\ State} 
    &  \makecell[l]{user\_slot: \\
          * food\_pref: None \\
          * loc: San Jose \\
          sys\_slot: \\
          * open: None \\
          * price: None \\
          * default: None \\
          * parking: None \\} 
    & \makecell[l]{
        terminal(term) \\
        succ(usr\_slot, food\_pref) \\
        succ(food\_pref, loc) \\
        succ(loc, term) \\
        usr\_slots(usr\_slot) \\
        known(usr\_slot)}\hspace{1em}
        \makecell[l]{
        known(loc) \\
        unknown(food\_pref) \\
        unknown(open) \\
        unknown(price) \\
        unknown(default) \\
        unknown(parking) \\} \\
    \hline
    \makecell[l]{User \\ Action} 
    &  \makecell[l]{
          act: inform \\
          param: loc}  
    & \makecell[l]{inform(loc)} \\
    \hline
    \makecell[l]{System \\ Action} &
    \makecell[l]{
          act: request \\
          param: food\_pref} 
    & \makecell[l]{sys\_request(food\_pref)}\\
    \hline
\end{tabular}

}
\end{minipage}
}
\end{table}

To adapt the SimDial problem to \name, 
the delexicalized \footnote{Delexicalization \cite{Henderson2014RobustDS} has be used previously to improve generalization in order to reinforce that the policy chooses a reasonable target.} state and actions are converted into the form of atoms, whose predicate is the action or state, and term is the slot. For example, $\mathtt{request}(${\it loc}$)$ denotes the action of requesting the location, while $\mathtt{unknown}(${\it price}$)$ indicates that the price is unknown to the system. Each turn is converted to a sample  $s$, where the background knowledge $\mathcal{B}$ is the combination of user actions and the belief state, and the positive examples $\mathcal{P}$ are the system actions. \name learns a mapping (a set of clauses) from the background  $\mathcal{B}$ to the positive examples $\mathcal{P}$. \edit{See Table~\ref{tab:illus} for an illustrative example of the adaption steps. The detailed process is described in Appendix~\ref{app:simdial}.}

Note that the four domains have different slots, and during training time, the model is only aware of the slots in the training set. During test time, the learned rules are directly applied to the converted samples in the test set. Additionally, to demonstrate continual learning, we add the pre-trained basic relationships of $\mathtt{all}$ and $\mathtt{member}$ to the background knowledge. $\mathtt{all}$ identifies that all items in a list satisfy some property, and $\mathtt{member}$ indicates that an item belongs to a list. See Appendix~\ref{app:simdial} for a more detailed description of the training process.

We compare \name with two baselines, the first one is Zero-Shot Dialog Generation (ZSDG) \cite{zhao2018zero}, which learns a cross-domain embedding of actions to enable the policy to zero-shot transfer to new domains. The second one is a naive multi-layer perceptron (MLP) model mapping from the encoding of the states to the actions. The models are trained and evaluated on the same datasets. We employ two metrics to quantify the performance of different models: {\bf Intent F1} measures whether the predicted dialog intent matches the ground truth, while {\bf Entity F1} measures whether the entity is predicted correctly. 

\begin{figure}[t]

\centering
    \caption{The evaluation results on the test set of SimDial comparing \name with MLP and ZSDG~\cite{zhao2018zero}. The In-Domain stands for the performance on the restaurant domain (500 samples, {\bf standard error: $\bf \pm 2.2 \%$}). Out-of-Domain one is the average performance on movie, bus, and weather (1500 samples, {\bf standard error: $\bf \pm 1.3 \%$}). See Table~\ref{tab:simdial_full} (appendix) for the per-domain results. The plot shows the zero-shot performance on movie and weather. -1000 denotes that the model is trained with 1000 dialogs from the restaurant domain. }
    \label{tab:simdial}
\begin{minipage}{0.53\textwidth}

\resizebox{\textwidth}{!}{

    \begin{tabular}{l|rr|rr}
    \hline
    &\multicolumn{2}{c}{In-Domain} & \multicolumn{2}{c}{Out-of-Domain} \\
     &Intent F1 & Entity F1 &  Intent F1 &  Entity F1 \\
     \hline
    ZSDG      &           91.41 &           13.48 &         83.08 &          0.59 \\
    MLP       &           91.81 &           87.52 &         61.22 &          5.50 \\
    DILOG     &           \textbf{99.74} &           \textbf{99.78} &         \textbf{99.75} &         \textbf{99.81} \\
    \hline
    ZSDG-1000 &           {\bf 99.98} &           {\bf 99.25} &         83.16 &         65.63 \\
    MLP-1000  &           98.97 &           97.98 &         54.49 &          6.63 \\
    \hline
    \end{tabular}
}
\end{minipage}
\hspace{1em}
\begin{minipage}{0.35\textwidth}
\includegraphics[width=\textwidth]{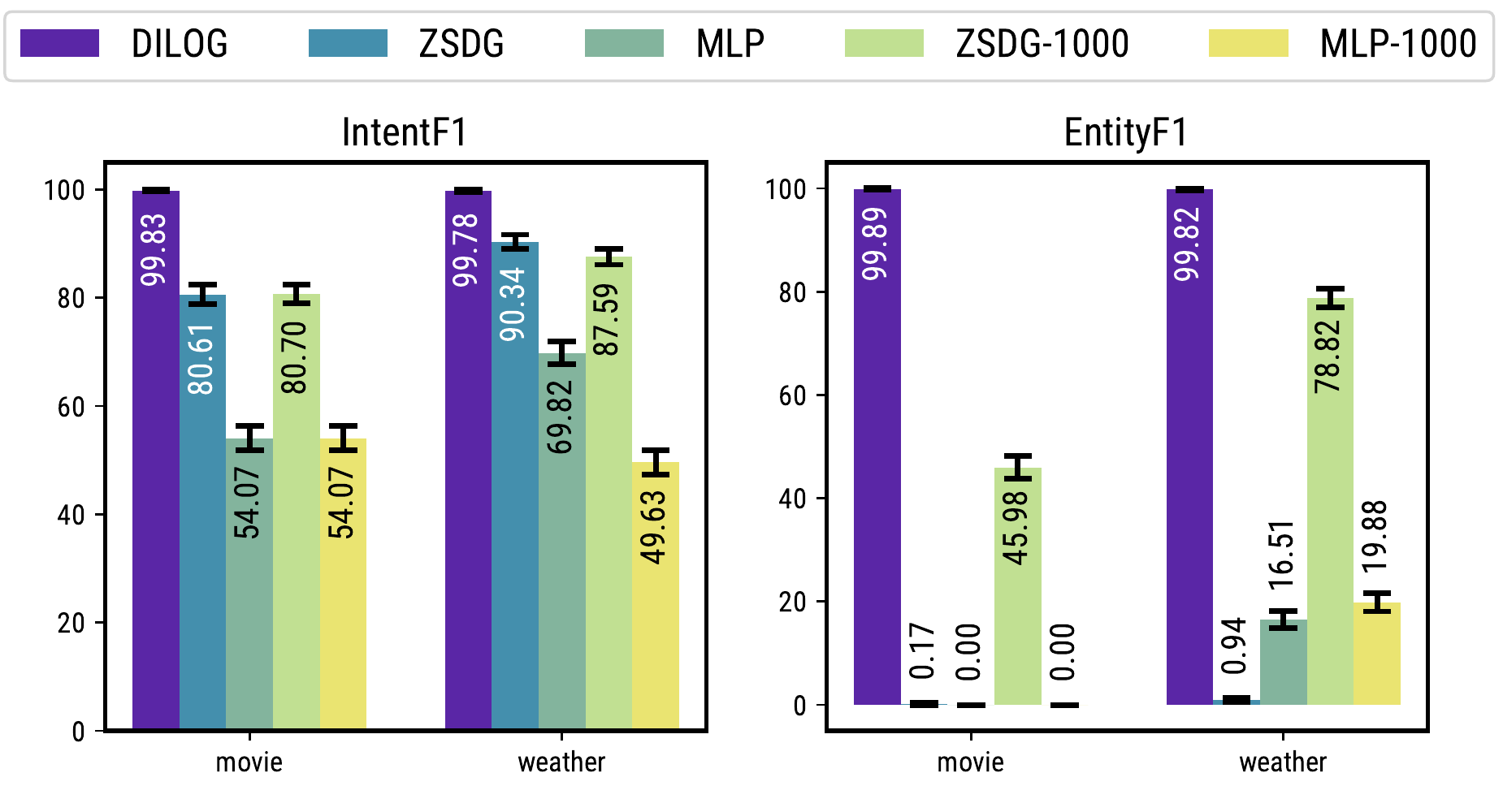}
\end{minipage}

\end{figure}

The evaluation results are shown in Figure~\ref{tab:simdial}. The In-Domain column demonstrates the performance of one-shot learning (trained with one dialog),
while the Out-of-Domain one shows the performance of zero-shot domain transfer. We also include the ZSDG and MLP models trained with 1000 samples (denoted with -1000) from the restaurant domain. On the SimDial dataset, \name consistently outperforms other models.

The better performance of \name on one-shot learning comes from the \emph{language bias} induced by the template that used to generate all the clauses, which can be regarded as a form of regularization. The ability of zero-shot domain transfer can be attributed to the \emph{symmetry} assumed by \name. For example the slots of $parking$ and $price$ are symmetrical in a sense that the rules applied to $parking$ should be directly applicable to $price$ as well. \name only breaks symmetry when necessary (for example, to differentiate between user slots and system slots), while maintaining the symmetry otherwise. However, in the vector-form encoding used by the neural networks, it is difficult, if not impossible, to express this symmetry.

The rules learned by \name can be extracted and interpreted by human beings. For example, the rule learned for when to request a slot is: 
   $$ \mathtt{sys\_request}(V0) \leftarrow \mathtt{member\_usr}(V0), \mathtt{unknown}(V0),$$
which reads: If a slot is one of the user slots, and that slot is unknown, the system should request that slot. The full set of learned rules are listed in  Appendix~\ref{app:learned_rules}. We also analyze an example that results in an error  in Appendix~\ref{app:error}, which is made possible by the interpretability of the \name framework.  
\ab{How are these rules related to the simulator used to generate the dialogs?}

% Recall that SimDial~\cite{zhao2018zero} is a simulated environment that generates dialogs based on a set of pre-defined rules. Hence, it is not surprising that DILP can learn a dialog policy by inferring all these underlying rules. Next, we explore the application of DILP to a more complex dataset.

\section{\name on the MultiWoZ Dataset}
\label{sec:multiwoz-zeroshot}

MultiWoZ~2.0 \cite{budzianowski2018multiwoz} is a large-scale human to human \edit {English}dialog dataset, which consists of dialogs on domains including restaurant, hotel, attraction, train, and taxi. It also includes dialogs that span multiple domains. The system response in MultiWoz dataset is stochastic, as human clerks have the freedom to choose from the possible actions. The actions are annotated by human annotators, which are noisy as well.
%\footnote{While MultiWoZ 2.1~\cite{eric2019multiwoz} aims at solving the noisy annotation issues, we choose to experiment on MultiWoZ 2.0 to quantify the impact of such noisy annotations on training of DILP.}
These features make MultiWoZ a more challenging task compared with SimDial. We used a variant of MultiWoZ~2.0 provided by ConvLab \cite{lee2019convlab}, which has annotated user actions.

The process of adapting the MultiWoZ dataset to \name is similar to that of SimDial (please refer to  Appendix~\ref{app:multiwoz} for the details). For training \name, we selected a single representative dialog whose system action contains all possible intents (inform, request, offerbooked, nooffer) from the restaurant domain. For testing, we use the original test set split in MultiWoZ dataset. 

\begin{figure}[t]
\centering
\caption{The evaluation results on the test set of MultiWoZ: \name is compared with MLP and DAMD~\cite{zhang2019task}. Note that DAMD applied its own heuristics in data preprocessing, so the Act. F1 of DAMD is not directly comparable with others. DAMD with the same preprocessing steps as others has a worse performance (Table~\ref{tab:multiwoz_full}). The Inform and Success metrics are not affected by preprocessing and therefore comparable. All stands for the performance on the all domains in the test set (1000 samples, {\bf standard error: $\bf \pm 1.6 \%$}). Out-of-Domain is the average performance of hotel, attraction, train, and taxi (1480 samples, {\bf standard error: $\bf \pm 1.3 \%$}). See Table~\ref{tab:multiwoz_full} (appendix) for the per-domain results. The plot shows the performance on two example domains of hotel and attraction. -AM denotes adding the missing slots in NLG, and -526 denotes the model trained with all 526 dialogs from the restaurant domain.}
\label{tab:multiwoz}
\begin{minipage}{0.5\textwidth}

\resizebox{\textwidth}{!}{
    \begin{tabular}{l|rrr|rrr}
    \hline
    &\multicolumn{3}{c}{All} & \multicolumn{3}{c}{Out-of-Domain} \\
    & Inform & Success & Act. F1 & Inform & Success & Act. F1 \\
    \hline
    DAMD     &         35.3 &           6.0 &            4.60 &       64.37 &         8.26 &          0.00 \\
    MLP      &         51.2 &          26.0 &           15.59 &       70.50 &        25.62 &          9.52 \\
    DILOG    &         \textbf{91.5} &          \textbf{43.4} &           23.67 &       \textbf{95.92} &       \textbf{ 65.49} &         \textbf{23.54} \\
    \hline

    DAMD-526 &         77.8 &          27.7 &           15.20 &       87.56 &        30.38 &          0.00 \\
    MLP-526  &         44.2 &          18.7 &           \textbf{29.23} &       64.02 &        17.56 &         14.64 \\
    \hline
    MLP-AM   &         51.4 &          43.3 &           15.59 &       70.44 &        42.15 &          9.52 \\
    DILOG-AM &         91.4 &          90.2 &           23.67 &       95.60 &        91.24 &         23.54 \\
    \hline
    \end{tabular}
}
\end{minipage}
\begin{minipage}{0.492\textwidth}
\includegraphics[width=\textwidth]{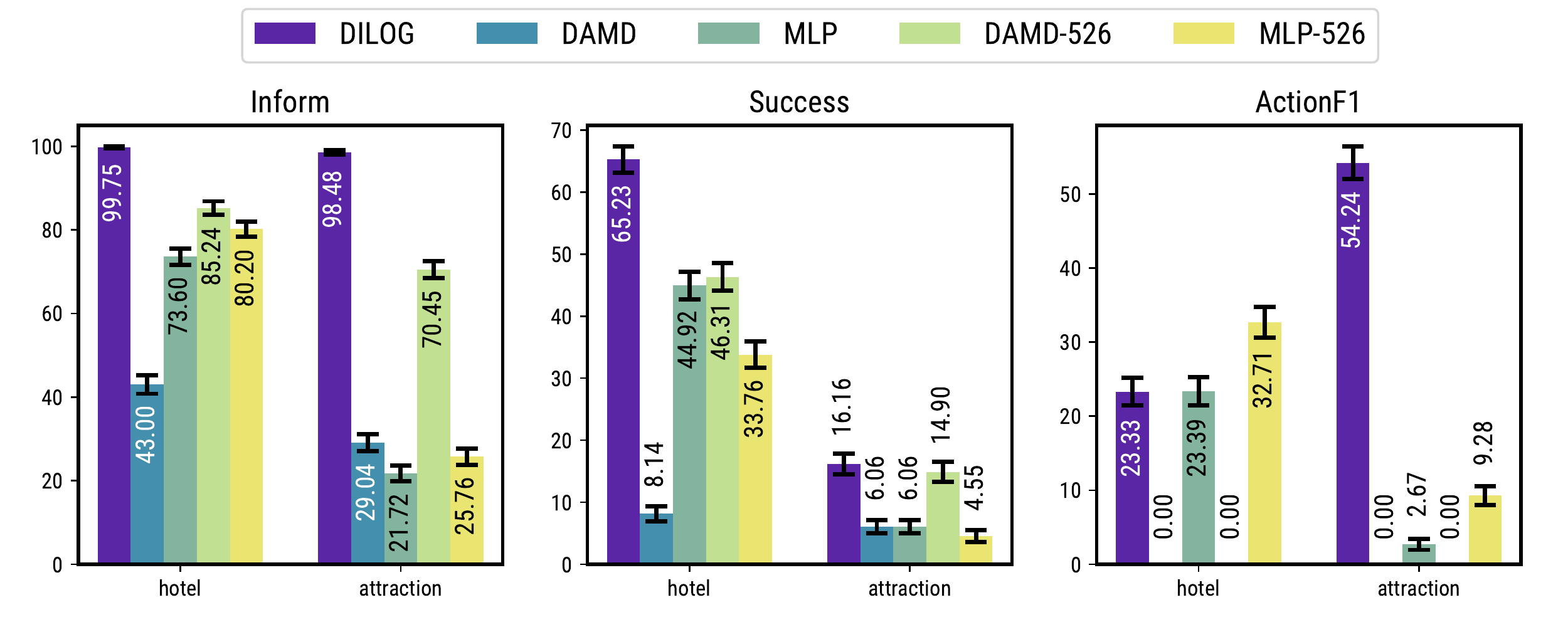}
\end{minipage}

\end{figure}

We compare \name with two baselines: DAMD \cite{zhang2019task} is the state-of-the-art model with dialog action prediction, and MLP is a naive baseline. We calculate the following metrics to evaluate the quality of the models: \emph{Inform} is a per diaog metric that measures whether the system finally provides a correct entity, \emph{Success} measures whether all the requested information are provided, and \emph{Action F1} is a per turn metric that checks if the predicted dialog action (both intent and entities) matches the ground truth. 
We use a template-based NLG to convert dialog actions predicted by \name and MLP to compute inform and success on generated utterances.  There are three slots (postcode, phone, address) that only appear in the system actions but not dialog states. We manually add an option to the template-based NLG to inform those slots whenever an entity is informed, and denote that variant as -AM, which significantly improves the success rate. We do not compare the -AM variant against other methods.

The results are shown in Figure~\ref{tab:multiwoz}. As can be seen, \name outperforms other models on the overall test set, which shows \name is capable of doing one-shot learning and zero-shot domain transfer even under noisy data. Compared with the MLP policy trained with all the 526 samples in the restaurant domain (denoted as -526), \name trained with one sample has a higher inform/success rate but lower action F1 score. Noticeably, \name with the addition of missing slots in NLG achieves 91.40 and 90.20 inform and success rate overall, which is higher than the state-of-the-art DAMD.\footnote{\edit{Note that DAMD is a more complex model with a higher capacity, which tends to overfit on extremely small datasets. Hence, MLP outperforms DAMD under the one-shot learning setting. However, when trained using all the dialogs in the restaurant domain, DAMD significantly outperformed the MLP when tested in-domain, as expected.} Also, we are training DAMD-526 solely on the 526 dialogs in Restaurant domain. This results in lower inform and success rates compared to DAMD trained on the entire MultiWoZ dataset, which are 89.2 and 77.9, respectively~\cite{zhang2019task}. } This shows inform and success are incomplete metrics which do not penalize false positives. Hence, we add action F1 as an additional metric to complement inform and success. Note that action F1 cannot solely capture the performance either, since there may be multiple possible true actions in a given state.

% \subsection{Few-Shot DILP Dialog Policy Learning with Few-Shot Transfer on MultiWoZ Dataset}
% \label{sec:multiwoz-fewshot}

% Experiment settings

% Results

% Conclude that these are competitive with state-of-the-art

\section{Conclusion}
\label{sec:conclusion}

In this paper, we introduce \name for resource constrained dialog policy learning and zero-shot domain transfer. Empirically, we demonstrate that \name outperforms strong neural baselines on SimDial and MultiWoZ datasets, while offering interpretability. We also provide an intuitive explanation on why \name shows these features. On the other hand, the \name framework has certain weaknesses, one being that it is computationally expensive when the template space gets large, where distributed training will be desired. Further, the program template used to generate all possible clauses needs to be hand-crafted, which is not a straightforward process. Another disadvantage is that real-valued inputs, such as confidence scores cannot be taken into account automatically. %It is possible to do pre-training to initialize the grounding atoms of $\mathtt{certain}()$ and $\mathtt{uncertain}()$ with the predicted value, but it need extra labels. 
The future work would naturally focus on solving these shortcomings. One way might be to jointly, in a multi-task setting, predict the templates or meta-learn them~\cite{minervini2020learning}.

\newpage

\section*{Acknowledgements}
The authors are thankful to Edward Grefenstette, Seungwhan Moon, Chinnadhurai Sankar, and Zhiguang Wang for constructive discussions about this work and the future directions, as well as Cristine Cooper, Drea Modugno, Sheeva Slovan, Heidi Young, and Kevin Ahlstrom for their feedback.
\bibliographystyle{coling}
\bibliography{ILP}
\newpage
\appendices
\setcounter{table}{0}
\renewcommand{\thetable}{A\arabic{table}}

\edit{
\section{Authoring the Program Template}
\label{app:progtemp}
A program template needs to be given to a ILP system, in order for it to generate the suitable rules. It is a general part of a system that does not need to be designed for each domain. Manual tuning is often required to get the desired program template. For example, a rule template is one of the important hyperparameters in the program template. A rule template is defined by the number of existentially quantified variables $v\in\mathbb{N} $ and whether to allow intensional predicates $i\in \{0, 1\}$. An \emph{existentially} quantified variable is a variable which is interpreted as "there exists ...". An \emph{intensional} predicate is a predicate defined by a set of clauses. Intuitively, the larger $v$, the more candidate rules will be generated, and thus easier for the model to overfit. Similarly, allowing intensional predicates will increase the number of rules generated. The hyperparameters can then be manually tuned to acheive the best performance. 

Alternatively, we can define the complexity of the rule template to be the number of candidate clauses generated under this rule template. One can order all the possible rule templates by their complexities, and do iterative search starting from the least complex rule template. It is also possible to perform a blackbox optimization on the hyperparameters in order to save the computation.

}

\section{Learning the Relationship of \texttt{all}}
\label{app:all}

The following example is used to learn the \texttt{all} relationship, which stands for all the items in a list satisfy some property.

The background knowledge $\mathcal{B}$ is

\begin{verbatim}
    terminal(t)
    succ(a, b)
    succ(b, c)
    succ(c, d)
    succ(d, e)
    succ(e, t)
    succ(f, g)
    succ(g, h)
    succ(h, t)
    true(a)
    true(c)
    true(d)
    true(e)
    true(f)
    true(g)
\end{verbatim}

The list is presented in a format of linked nodes, where $\mathtt{succ}(A, B)$ means the successor of $A$ is $B$,  $\mathtt{terminal}(A)$ means $A$ is a terminal state, and $\mathtt{true}(A)$ specifies that the property under consideration holds for item $A$. In this example, the property does not hold for $b$ and $h$.  The positive examples are 

\begin{verbatim}
    all(c), all(d), all(e)
\end{verbatim}

and the negative examples are 
\begin{verbatim}
    all(a), all(b), all(f), all(g), all(h)
\end{verbatim}

The learned clauses for \texttt{all} are

\begin{verbatim}
    pred1(V0, V1) <- succ(V0, V1), all(V1)
    pred1(V0, V1) <- succ(V0, V1), terminal(V1)
    all(V0) <- true(V0), pred1(V0, V1)
\end{verbatim}

\section{Preprocessing Details for SimDial}
\label{app:simdial}

\subsection{Data Conversion}

We illustrate the data conversion process using the following dialog:

\begin{lstlisting}
  SYS: Welcome to restaurant recommendation system. How can I help?
  USR: What's up? I need a restaurant.
  SYS: Which place?
  USR: I am at San Jose.
  SYS: What kind of food do you like?
  USR: Chinese.
  SYS: {"QUERY": {"#loc": "San Jose", "#food_pref": "Chinese"}, "GOALS": ["#default"]}
  USR: {"RET": {"#default": "6"}}
  SYS: Restaurant 6 is a good choice. Are you happy about my answer?
  USR: One more thing. Tell me if the restaurant is open.
  SYS: {"QUERY": {"#loc": "San Jose", "#food_pref": "Chinese"}, "GOALS": ["#open"]}
  USR: {"RET": {"#open": "open"}}
  SYS: The restaurant is open. Are you happy about my answer?
  USR: I have more requests. How expensive it is?
  SYS: {"QUERY": {"#loc": "San Jose", "#food_pref": "Chinese"}, "GOALS": ["#price"]}
  USR: {"RET": {"#price": "expensive"}}
  SYS: The restaurant serves expensive food. Anything else?
  USR: All good. That's all.
  SYS: Goodbye.
\end{lstlisting}

To adapt the DILP framework to the SimDial problem, the delexicalized states are converted to the form of atoms. For example, the state of the second turn is

\begin{verbatim}
'usr_slots': [{'name': '#loc', 'max_conf': 0.99, 'max_val': 'San Jose'},
              {'name': '#food_pref', 'max_conf': 0.0, 'max_val': None}],
'sys_goals': [{'name': '#default','delivered': False,'value': None,
               'expected': None, 'conf': 1.2},
              {'name': '#open','delivered': False,'value': None,
               'expected': None, 'conf': 0.0},
              {'name': '#price', 'delivered': False, 'value': None,
               'expected': None, 'conf': 0.0},
              {'name': '#parking', 'delivered': False, 'value': None,
               'expected': None, 'conf': 0.0}]
\end{verbatim}

It was converted to the following grounding atoms:

\begin{verbatim}
    known(loc)
    unknown(food_pref)
    terminal(term)
    known(usr_slot)
    usr_slots(usr_slot)
    succ(usr_slot, food_pref)
    succ(food_pref, loc)
    succ(loc, term)
    unknown(default)
    unknown(open)
    unknown(price)
    unknown(parking)
\end{verbatim}

Note that we use a linked list to enumerate the user slots, where $\mathtt{succ}(A, B)$ means the successor of $A$ is $B$, and $\mathtt{terminal}(A)$ means $A$ is a terminal state.

User and system actions are converted to the form of atoms in a similar way, for example, the user action of 

\begin{verbatim}
    [{'act': 'inform', 'parameters': [['#loc', 2]]}]
\end{verbatim}

is converted to $\mathtt{inform}(loc)$, and the system action of 

\begin{verbatim}
    [{'act': 'request', 'parameters': [['#food_pref', None]]}]
\end{verbatim}

is converted to $\mathtt{sys\_request}(food\_pref)$.

The constants $\mathcal{C}$ are all the slot values appeared in the state and actions:

\begin{verbatim}
    loc
    food_pref
    default
    open
    price
    parking
    term
    usr_slot
\end{verbatim}

At each turn, the user action and state are combined to be the background knowledge $\mathcal{B}$, the constants $\mathcal{C}$ are all the possible slots, the positive examples $\mathcal{P}$ are the system actions, and the negative examples $\mathcal{N}$ are all other actions.

\subsection{Domain Transfer}

The model induced from the restaurant domain is tested on the other three domains of movie, weather and bus. The user slots and system goal slots can be different. For example, in the movie domain, the slots are 

\begin{verbatim}
    genre
    years
    default
    country
    rating
    company
    director
    term
    usr_slot
\end{verbatim}

We can still follow the same conversion procedure for the restaurant domain. For example, the background $\mathcal{B}$ of 

\begin{verbatim}
    known(genre)
    known(years)
    unknown(country)
    terminal(term)
    known(usr_slot)
    usr_slots(usr_slot)
    succ(usr_slot, genre)
    succ(genre, years)
    succ(years, country)
    succ(country, term)
    unknown(default)
    unknown(rating)
    unknown(company)
    unknown(director)
    inform(years)
    request(default)
\end{verbatim}

can be mapped to the positive example of $\mathtt{sys\_request}(country)$.

\subsection{Results}

The evaluation results on the SimDial dataset for different domains are shown in Table~\ref{tab:simdial_full}. \edit{The performance of different domains are similar.}

\begin{table}[H]
    \centering
    \scalebox{0.9}{
    \begin{tabular}{c|cc|cc|cc|cc}
    \hline
         & \multicolumn{2}{c}{restaurant} & \multicolumn{2}{c}{movie}  & \multicolumn{2}{c}{bus}  &  \multicolumn{2}{c}{weather}  \\
         & Intent F1 & Entity F1 & Intent F1 & Entity F1 & Intent F1 & Entity F1 & Intent F1 & Entity F1\\
    \hline
    ZSDG & 91.41 & 13.48 & 80.61 & 0.17 & 78.30 & 0.65 & 90.34 & 0.94  \\
    MLP  & 91.81 & 87.52 & 54.07 & 0.0 & 59.78 & 0.0 & 69.82 & 16.51 \\
    \name & \textbf{99.74} & \textbf{99.78} & \textbf{99.83} & \textbf{99.89} &  \textbf{99.63} & \textbf{99.72} & \textbf{99.78} & \textbf{99.82} \\
    \hline
    ZSDG-1000 & 99.98 & 99.25 & 80.70 & 45.98 & 81.20 & 72.09 & 87.59 & 78.82 \\
    MLP-1000 & 98.97 & 97.98 & 54.07 & 0.00 & 59.78 & 0.00 & 49.63 & 19.88 \\
    \hline
    \end{tabular}
    }
    \caption{\small The evaluation results on the test set of SimDial, \name is compared with MLP and ZSDG~\cite{zhao2018zero}. The models are trained on a dialog from the restaurant doamin and evaluated on all other domains. We also included the ZSDG and MLP model trained with 1000 samples from the restaurant domain as an upper bound.
    \ab{Can we report standard errors for these numbers? I understand that you are not going to be training these models multiple times. Say if the computed number here is $x = 1/N \sum_{i} s_i$ where $s_i \in \{0, 1\}$, then we can compute $\sigma_x = \sqrt{x (1-x)/N}$ as a metric showing how much variation there is due to the finite size of the test set. This would not capture the variations in training data, and other randomness that might be there in training your model, but since the training data is chosen to be the same for all methods, it could still be a reasonable surrogate on how to compare the numbers we are achieving from the different models.
} \ab{What caused the 0.2\% performance drop? Can we have an example for the error? Is it a rare rule that we are missing?} \ab{Can we have a few full dialog examples in the appendix?} \ab{Can we also compare to ZSDG with X shot transfer and increase X so that ZSDG matches the performance of DILP and report that X? Would it be 10-100-1000? That could strengthen your argument.}}
    \label{tab:simdial_full}
\end{table}
\begin{table}[h]
\small
    \centering
    \begin{tabular}{l|l|l}
    \hline
    Interaction & True Action & Predicted Action \\
    \hline
    \begin{lstlisting}
USR: Hi. I need a restaurant.
SYS: Which city are you interested in?
    \end{lstlisting}
 &     \begin{lstlisting}
sys_request(loc)
sys_request(food_pref)
    \end{lstlisting} 
         & \begin{lstlisting}
sys_request(loc)
sys_request(food_pref)
    \end{lstlisting} \\ 
    \hline
    
    \begin{lstlisting}
USR: I am at Boston.
SYS: What type of restaurant?
    \end{lstlisting}
 & sys\_request(food\_pref)
 & sys\_request(food\_pref)  \\ 
    \hline
    
        \begin{lstlisting}
USR: French restaurant.
SYS: {"QUERY": {"#loc": "Boston", 
                "#food_pref": "French"},
      "GOALS": ["#default"]}
    \end{lstlisting}
 & sys\_query(default)
 & sys\_query(default)  \\ 
    \hline
    
            \begin{lstlisting}
USR: {"RET": {"#default": "56"}}
SYS: Restaurant 56 is a good choice. Anything else?
    \end{lstlisting}
 & sys\_inform(default)
 & sys\_inform(default)  \\ 
    \hline
    
    \begin{lstlisting}
USR: I like Chinese food.
SYS: {"QUERY": {"#loc": "Boston", 
               "#food_pref": "Chinese"}, 
      "GOALS": ["#default"]}
    \end{lstlisting}
 & sys\_query(default)
 & {\color{red} None}  \\ 
    \hline

   \begin{lstlisting}
USR: {"RET": {"#default": "94"}}
SYS: Restaurant 94 is a good choice. What else can I do?
    \end{lstlisting}
 & sys\_inform(default)
 & sys\_inform(default)  \\ 
    \hline
    
       \begin{lstlisting}
 USR: One more thing. How easy is it to park?
  SYS: {"QUERY": {"#loc": "Boston", 
                 "#food_pref": "Chinese"},
        "GOALS": ["#parking"]}
    \end{lstlisting}
 & sys\_query(parking)
 & sys\_query(parking)  \\ 
    \hline
    
   \begin{lstlisting}
USR: {"RET": {"#parking": "valet parking"}}
SYS: The restaurant has valet parking. What else can I do?
    \end{lstlisting}
 & sys\_inform(parking)
 & sys\_inform(parking)  \\ 
    \hline

     \begin{lstlisting}
USR: One more thing. How expensive it is?
SYS: {"QUERY": {"#loc": "Boston", 
                  "#food_pref": "Chinese"}, 
        "GOALS": ["#price"]}

    \end{lstlisting}
 & sys\_query(price)
 & sys\_query(price)  \\ 
    \hline
    
   \begin{lstlisting}
USR: {"RET": {"#price": "expensive"}}
SYS: The restaurant serves expensive food. 
       Anything else?
    \end{lstlisting}
 & sys\_inform(price)
 & sys\_inform(price)  \\ 
    \hline
    
    USR: No more questions. That's all & & \\
    \hline

    \end{tabular}
    \caption{An example that DILOG predicted a wrong dialog action.}
    \label{tab:failure_example}
\end{table}

\subsection{Learned rules}
\label{app:learned_rules}

The rules learned by DILP are simple and can be extracted and interpreted by humans. In particular, DILP converges to these set of rules with  probabilities close to one except for the predicate of \texttt{sys\_inform}:
\begin{align*}
    \mathtt{sys\_request}(V0) &\leftarrow \mathtt{member\_usr}(V0), \mathtt{unknown}(V0),\\
    \mathtt{sys\_inform}(V0) &\leftarrow \mathtt{kb\_return}(V0), \\
    \mathtt{sys\_query}(V0) &\leftarrow \mathtt{request}(V0), \mathtt{pred3}(V0),\\
    \mathtt{pred2}() &\leftarrow \mathtt{all}(V0), \mathtt{usr\_slots}(V0),\\
    \mathtt{pred3}(V0) &\leftarrow \mathtt{pred2}(), \mathtt{unknown}(V0),
\end{align*}
where $\mathtt{pred2}$ and $\mathtt{pred3}$ are {\it invented} predicates used as intermediate states. As an example, the first rule reads: If a slot is one of the user slots, and that slot is unknown, the system should request that slot.

\subsection{A failure case}
\label{app:error}

While \name achieves 99+\% Intent F1 and Entity F1 scores, we wanted to understand the rare cases that lead to failures.
A lemon-picked dialog in which \name failed to predict the correct system action is listed in Table~\ref{tab:failure_example}.

% \begin{lstlisting}
%   SYS: Welcome to restaurant recommendation system. How can I help?
%   USR: Hi. I need a restaurant.
%   SYS: Which city are you interested in?
%   USR: I am at Boston.
%   SYS: What type of restaurant?
%   USR: French restaurant.
%   SYS: {"QUERY": {"#loc": "Boston", "#food_pref": "French"}, "GOALS": ["#default"]}
%   USR: {"RET": {"#default": "56"}}
%   SYS: Restaurant 56 is a good choice. Anything else?
%   USR: I like Chinese food.
%   SYS: {"QUERY": {"#loc": "Boston", "#food_pref": "Chinese"}, "GOALS": ["#default"]}
%   USR: {"RET": {"#default": "94"}}
%   SYS: Restaurant 94 is a good choice. What else can I do?
%   USR: One more thing. How easy is it to park?
%   SYS: {"QUERY": {"#loc": "Boston", "#food_pref": "Chinese"}, "GOALS": ["#parking"]}
%   USR: {"RET": {"#parking": "valet parking"}}
%   SYS: The restaurant has valet parking. What else can I do?
%   USR: One more thing. How expensive it is?
%   SYS: {"QUERY": {"#loc": "Boston", "#food_pref": "Chinese"}, "GOALS": ["#price"]}
%   USR: {"RET": {"#price": "expensive"}}
%   SYS: The restaurant serves expensive food. Anything else?
%   USR: No more questions. That's all.
%   SYS: Goodbye.
% \end{lstlisting}

The error happens in the 5th turn when user corrected the food\_pref slot he previously provided. The true system action should be $\mathtt{sys\_query}(default)$, while the predicted one is \texttt{None}, which means no action should be taken. We can also pin-point the erroneous rule to be $\mathtt{pred3}$. All these error analysis are made possible by the explainable rules from the DILP framework. By fixing these errors (retraining on scenarios where such slot correction is provided), we expect \name to fully learn SimDial with 100\% Intent F1 and Entity F1 scores. However, that is not the point of this study.

\section{Preprocessing Details for MultiWoZ}
\label{app:multiwoz}

We illustrated the data conversion process for MultiWoZ using the following dialog:

\begin{lstlisting}
  USR: I am looking to get some eritrean food in the west part of town.
  SYS: Unfortunately there aren't any restaurants that fit your criteria. 
       Is there anything else that I can do for you?
  USR: Can you see if there are any that serve indian food?
  SYS: There are six restaurants, one in the moderate price range and 
       the rest in the expensive price range. 
       Which one would you prefer?
  USR: Any would be fine. But I need to book a table for 8 people for Sunday at 12:15.
  SYS: Booking was unsuccessful at rice boat. Would you like to find another restaurant?
  USR: Sure as long as it's in the same area and price range. I need the reference number too please.
  SYS: Okay I booked you at the Tandoori Palace and your reference number is NBLD7E0V.
  USR: Thank you, that's all I needed today!
  SYS: Okay glad I could be of help!
\end{lstlisting}

In the first turn, the state of restaurant

\begin{verbatim}
   'restaurant': 
        {'book': {'booked': [], 'people': '', 'day': '', 'time': ''},
         'semi': {'food': 'eritrean',
                  'pricerange': 'not mentioned',
                  'name': 'not mentioned',
                  'area': 'west'}}
\end{verbatim}

is converted to the following grounding atoms:

\begin{verbatim}
    "usr_inform(food)",
    "usr_inform(area)",
    "known(food)",
    "unknown(price)",
    "unknown(name)",
    "known(area)",
    "unknown(people)",
    "unknown(day)",
    "unknown(time)"
\end{verbatim}

We also included the information of whether the book or match is successful from the database pointers. When there is no match or booking failed, $\mathtt{no\_match}()$ or $\mathtt{book\_fail}()$ is added to the state as well.

In the conversion of actions, we ignored all actions in the general domain including greeting, bye, thanks, etc. Besides, the system actions of select, recommend, and offerbook are treated to be the same as inform, as they are all providing information to the user. For example, the user actions of 

\begin{verbatim}
    [['inform', 'restaurant', 'food'],
     ['inform', 'restaurant', 'area']]
\end{verbatim}

are converted to the atoms of $\mathtt{inform}(food)$ and $\mathtt{inform}(area)$, while the system actions of 
\begin{verbatim}
    [['nooffer', 'restaurant', 'none'],
     ['reqmore', 'general', 'none']]
\end{verbatim}

are converted to the atoms of $\mathtt{nooffer}()$.

The construction of the sample $(\mathcal{B}, \mathcal{P}, \mathcal{N}, \mathcal{C})$ is similar to that of SimDial. Note that during training, we ignored all the domain information. During inference time, we separate the belief state by domains, and run inference on each domain using the same model. Finally, the predicted actions for each domain are combined to yield the final action prediction.

\subsection{Results}

The evaluation results for different domains on the MultiWoZ dataset are shown in Table~\ref{tab:multiwoz_full}. \edit{The performance of different domains are similar except train and taxi, this is because the goals in those two domains are less diverse.} DAMD applied its own heuristics in data preprocessing, so the Act. F1 of DAMD is not directly comparable with others. We also included DAMD with the same data-preprocessing as the others (denoted as DAMD'), whose performance is worse.

\begin{table}[H]
%\hspace*{-4pt}
\setlength{\tabcolsep}{2pt}
\scriptsize
    \centering
   % \hspace*{-3.5em}
    \begin{tabular}{c%|ccc
    |ccc|ccc|ccc|ccc|ccc}
    \hline
         & %\multicolumn{3}{c}{all}& 
         \multicolumn{3}{c}{restaurant} & \multicolumn{3}{c}{hotel} & \multicolumn{3}{c}{attraction} & \multicolumn{3}{c}{train} & \multicolumn{3}{c}{taxi} \\
         %& inform & success & action F1
         & inform & success & action F1 & inform & success & action F1 & inform & success & action F1 & inform & success & action F1 & inform & success & action F1 \\
    \hline
    DAMD & %35.3 & 6.0 & 4.6 & 
    50.11 & 9.15 & 2.81 & 43.0 & 8.14 & 0.00 & 29.04 & 6.06 & 0.00 & 85.43 & 18.83 & 0.00 & 100.0 & 0.00 & 0.00 \\
    MLP  & %51.20 & 26.00 & 15.59 &
    83.30 & 27.92 & 23.74& 73.60 & 44.92 & \textbf{23.39} & 21.72 & 6.06 & 2.67 & \textbf{86.67} & 51.52 &12.01& 100.0 & 0.00 & 0.00\\
    \name & %\textbf{91.50} & \textbf{43.40} & \textbf{23.67} &
    \textbf{98.17} & \textbf{60.41} &\textbf{27.62} & \textbf{99.75} & \textbf{65.23} & 23.33 & \textbf{98.48} & \textbf{16.16} & \textbf{54.24} & 85.45 & \textbf{84.65} &\textbf{16.26}& 100.0 & \textbf{95.90} & \textbf{0.33}\\
    \hline
    MLP-AM  & %51.40 & 43.30 & 15.59 & 
    84.44 & 40.05 & 23.74& 73.35 & 52.28 & 23.39 & 21.72 & 17.42 & 2.67 & 86.67 & 51.72 & 12.01 & 100.0 & 47.18 & 0.00 \\
    \name-AM & %91.40 & 90.20 & 23.67 &
    99.08 & 91.30 & 27.62 & 98.98 & 93.65 & 23.33 & 97.98 & 90.15 & 54.24 & 85.45 & 84.24 & 16.26& 100.0 & 96.92 & 0.33\\
    \hline
    DAMD-526 & %77.8 & 27.7 & 15.2 &
    93.59 & 51.72 & 13.65 & 85.24 & 46.31 & 0.00 & 70.45 & 14.90 & 0.00 & 94.53 & 60.32 & 0.00 & 100.0 & 0.00 & 0.00 \\
    MLP-526 & %44.20 & 18.70 & 29.23 &
    89.02 & 31.12 & 52.83 & 80.20 & 33.76 & 32.71  & 25.76 & 4.55 & 9.28 & 50.10 & 31.92 & 16.56 & 100.0 & 0.00 & 0.00 \\
    \hline
    DAMD' & 21.97 & 1.83 & 0.00 & 29.70 & 3.05 & 0.00 & 23.74 & 4.04 & 0.00 & 85.66 & 10.91 & 0.00 & 100.0 & 0.00 & 0.00 \\
    DAMD'-526 & 92.22 & 22.88 & 14.02 & 88.58 & 19.54 & 0.00 & 68.70 & 14.14 & 0.00 & 94.75 & 40.20 & 0.00 & 100.0 & 1.03 & 0.00 \\
    \hline
    \end{tabular}
    \caption{The evaluation results on the test set of MultiWoZ, \name is compared with MLP and DAMD~\cite{zhang2019task}. The models are trained on a dialog from the restaurant domain and evaluated on all other domains. We also included the DAMD and MLP model trained with all 526 samples from the restaurant domain as an upper bound. \ab{Can we think of breaking these results into a table and a figure? Maybe we keep restaurant and all in one table and then create a plot for the different domains? Also we need to report the standard errors for all reported numbers to ensure the results are statistically significant.}
    \ab{When it comes to results, it would also be nice to report {\ab inference time} comparisons with other baselines to pitch resource constraints.} \ab{I think we also need DAMD 1000. In particular, what would be DAMD 1000's action F1? and Can you match that? Maybe we should also try MLP 1000 as well.} \ab{It would be nice to also compare with few (X) shot transfer learning and conclude that your zero-shot transfer performs similarly to DAMD's X shot transfer.}\ab{Can we include a few examples of trained policies in the appendix?}}
    \label{tab:multiwoz_full}
\end{table}

\end{document}